%% file: iccad25.tex
\documentclass[10pt,conference]{IEEEtran}
\IEEEoverridecommandlockouts 
\input{setting-ieee}

\usepackage{amsmath}   
\usepackage{graphicx}   
\usepackage{adjustbox} 
\usepackage{pgf-pie}
\usepackage{tikz}
\usetikzlibrary{calc}

\definecolor{lightcyan}{RGB}{162, 221, 208}
\definecolor{darkblue}{RGB}{70, 130, 180}
\definecolor{lightblue}{RGB}{173, 216, 230}
\definecolor{lightyellow}{RGB}{255, 255, 204}
\definecolor{orange}{RGB}{255, 165, 0}
\definecolor{lightgreen}{RGB}{144, 238, 144}
\definecolor{lightgray}{RGB}{211, 211, 211}       
\definecolor{myblue}{RGB}{154, 164, 208}
\definecolor{myred}{RGB}{242, 200, 195}
\definecolor{mypurple}{RGB}{197, 176, 213}

\definecolor{myblue}{RGB}{29,114,221}    
\definecolor{myyellow}{RGB}{255,255,191} 
\definecolor{myorange}{RGB}{244,106,18}  
\definecolor{mygray}{RGB}{102,102,102}   
\definecolor{mypink}{RGB}{252,228,215}   

\definecolor{CUpurple}{RGB}{117,15,109}
\definecolor{CUlpurple}{RGB}{165,133,182}
\definecolor{CUgold}{RGB}{221,163,0}
\definecolor{CUribbon}{RGB}{244,223,176}
\definecolor{CUblack}{RGB}{34,24,21}
\definecolor{PKUred}{RGB}{126,24,28}
\definecolor{gray6}{gray}{0.6}
\definecolor{gray7}{gray}{0.7}
\definecolor{gray8}{gray}{0.8}
\definecolor{gray9}{gray}{0.9}

\graphicspath{{./figs/}{./submission/figs/}}

\begin{document}
\date{}

\title{
    Invited Paper: Unitho: A Unified Multi-Task Framework for Computational Lithography
}

\author{%
    \IEEEauthorblockN{Qian Jin$^{1,\ \dagger}$, Yumeng Liu$^{1,\ \dagger}$, Yuqi Jiang$^{1,\ \dagger}$, Qi Sun$^{1,\ *}$, Cheng Zhuo$^{1,\ *}$}
    \IEEEauthorblockA{$^1$Zhejiang University, Hangzhou, China}\\
    \IEEEauthorblockA{$^\dagger$Equal Contributions; $^*$Corresponding Authors: \{qisunchn, czhuo\}@zju.edu.cn}
}

\maketitle

\pagestyle{plain}
\input{doc/abstract} 
\input{doc/intro} 
\input{doc/prelim}
\input{doc/algo}
\input{doc/result}

\input{doc/conclu}

\clearpage
{
    \bibliographystyle{IEEEtran}
    \bibliography{main}
}

\end{document}

%% file: setting-ieee.tex
\usepackage{blkarray}                                      
\usepackage{graphicx}                                      
\usepackage{amsmath}
\usepackage{amssymb}
\usepackage{amsfonts}
\usepackage{amsthm}
\usepackage[mathcal]{eucal}
\usepackage{mathrsfs}
\usepackage{booktabs}
\usepackage{enumerate}
\usepackage{multirow}
\usepackage[subrefformat=parens,farskip=0pt,justification=centering]{subfig}
\usepackage{color}
\usepackage{cite}                                          
\usepackage{comment}                                       
\usepackage{soul}                                          
\soulregister\cite7
\soulregister\ref7
\soulregister\pageref7
\usepackage{etoolbox}                                      
\usepackage{url}
\usepackage{nth}                                           
\usepackage{bm}                                            
\usepackage{courier}
\usepackage{balance}
\usepackage{threeparttable}
\usepackage{xcolor,colortbl}
\usepackage{footnote}
\usepackage{listings}
\usepackage{setspace}                                      
\usepackage[inline]{enumitem}
\usepackage{verbatim}
\usepackage[bookmarks=false]{hyperref}
\hypersetup{
    colorlinks = true,
    citecolor  = blue,
    linkcolor  = blue,
    urlcolor   = blue,
}
\usepackage{tikz}
\usetikzlibrary{patterns,snakes}
\usetikzlibrary{positioning,calc,fit,decorations.pathmorphing,shapes.geometric, shapes.gates.logic.US, calc}
\usetikzlibrary{arrows,arrows.meta,decorations.markings,shapes,shapes.arrows}
\usetikzlibrary{decorations,decorations.pathreplacing}
\usetikzlibrary{backgrounds}
\usepackage{filecontents}                                  
\usepackage{pgfplots}
\usepackage{pgfplotstable}
\usepackage{scalefnt}
\pgfplotsset{compat=newest}
\usepackage{caption}
\usepackage{pifont}                                        
\usepackage{cleveref}
\Crefformat{figure}{Fig.~#2#1#3}                           
\Crefname{subfigure}{Fig.}{Figs.}
\Crefname{figure}{Fig.}{Figs.}
\Crefformat{table}{TABLE~#2#1#3}                           
\usepackage[figuresright]{rotating}

\usepackage{algorithm}
\iftrue
\usepackage[noend]{algpseudocode}
\algrenewcommand\textproc{\texttt}
\makeatletter
\let\OldStatex\Statex
\renewcommand{\Statex}[1][3]{%
  \setlength\@tempdima{\algorithmicindent}%
  \OldStatex\hskip\dimexpr#1\@tempdima\relax
}
\makeatother
\else
\usepackage{algorithmic}
\fi

\definecolor{CUHKorange}{RGB}{244,106,18} 
\definecolor{CUHKblue}{RGB}{0,111,190}    
\definecolor{CUHKgreen}{RGB}{0,127,128}   
\definecolor{CUHKred}{RGB}{228,46,36}     
\definecolor{CUHKyellow}{RGB}{198,148,34} 
\definecolor{CUHKdark}{RGB}{114,44,114}   
\definecolor{CUHKmiddle}{RGB}{144,44,144} 
\definecolor{CUHKlight}{RGB}{167,44,167} 
\definecolor{CUHKpurple}{RGB}{117,15,109}
\definecolor{CUHKgold}{RGB}{221,163,0}
\definecolor{CUHKribbon}{RGB}{244,223,176}
\definecolor{CUHKblack}{RGB}{34,24,21}

\renewcommand{\bf}[1]{\textbf{#1}}
\renewcommand{\tt}[1]{\texttt{#1}}

\newcommand{\minisection}[1]{\vspace{.06in}\noindent{\textbf{#1}}.}

\usepackage{tcolorbox}
\tcbuselibrary{skins,breakable}
    {\endtcolorbox}
    {\endtcolorbox}

\usepackage[top=0.80in,bottom=0.88in,left=0.63in,right=0.63in]{geometry}
\iftrue
\newcommand{\subparagraph}{}
\usepackage{titlesec}
\titlespacing*{\section}{0pt}{1.8ex plus .2ex minus .2ex}{0.4ex plus .2ex}
\titlespacing*{\subsection}{0pt}{1.0ex plus .2ex minus .2ex}{0.2ex plus .2ex}
\fi

\newtheorem{myproblem}{\textbf{Problem}}

\crefname{mytheorem}{Theorem}{Theorems}
\crefname{mylemma}{Lemma}{Lemmas}
\crefname{myclaim}{Claim}{Claims}
\crefname{myproperty}{Property}{Properties}
\crefname{mycorollary}{Corollary}{Corollaries}

\RequirePackage[normalem]{ulem} 
\RequirePackage{color}\definecolor{RED}{rgb}{1,0,0}\definecolor{BLUE}{rgb}{0,0,1} 

\newcommand{\todo}[1]{\textcolor{red}{[TODO: #1]}}

%% file: doc/abstract.tex
\begin{abstract}
Reliable, generalizable data foundations are critical for enabling large-scale models in computational lithography. However, essential tasks—mask generation, rule violation detection, and layout optimization—are often handled in isolation, hindered by scarce datasets and limited modeling approaches. To address these challenges, we introduce Unitho, a unified multi-task large vision model built upon the Transformer architecture. Trained on a large-scale industrial lithography simulation dataset with hundreds of thousands of cases, Unitho supports end-to-end mask generation, lithography simulation, and rule violation detection. By enabling agile and high-fidelity lithography simulation, Unitho further facilitates the construction of robust data foundations for intelligent EDA. Experimental results validate its effectiveness and generalizability, with performance substantially surpassing academic baselines.
\end{abstract}


%% file: doc/intro.tex
\section{Introduction}
\label{sec:introduction}

\begin{figure}[tb]
    \centering
    \includegraphics[width=\linewidth]{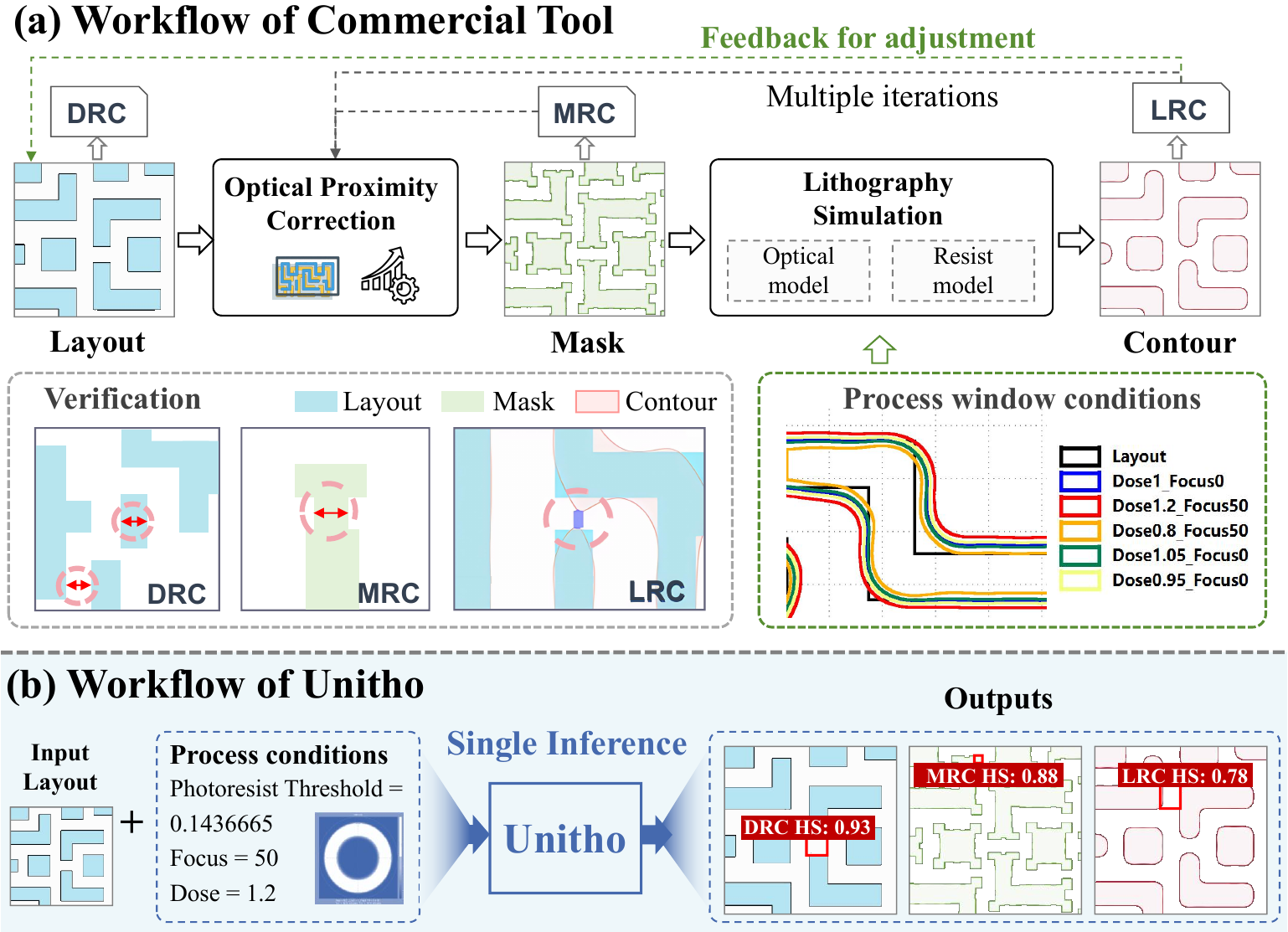}
    \caption{Workflow Comparison. (a) Traditional processing flow of commercial tools; (b) Unitho’s substitution role, accelerating iterations.}
    \label{fig:workflow}
\end{figure}

\begin{figure}[tb]
    \input{figs/runtime} 
    \caption{
    Runtime Comparison of Commercial Tool and Unitho on 6000 × 6000 nm layout clips. Time cost breaks into 3 stages: OPC, Lithography Simulation, and Verifications (DRC+MRC+LRC), with the commercial tool executed on a 20-core Intel Xeon Platinum server for 30 OPC iterations.}
    \label{fig:time-breakdown}
\end{figure}
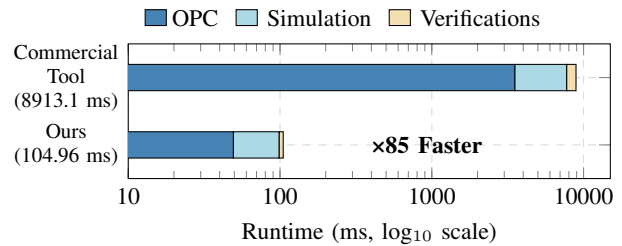

As process nodes continue to shrink, geometric distortions induced by photolithography, such as optical proximity effects (OPE), 
pose a growing challenge to device performance and manufacturing yield. To ensure that design layouts are transferred to the wafer with high fidelity, optical proximity correction (OPC) and subsequent lithography verification have become indispensable steps in the chip design workflow \cite{yang2022machine}. However, the industry-standard physics-based simulation, while accurate, is computationally intensive and time-consuming, as shown in \Cref{fig:time-breakdown}
This bottleneck is severely exacerbated during process window (PW) analysis, which requires validating design robustness under variations in focus and exposure dose. Since simulations must be repeated across the entire process matrix, the resulting computational overhead significantly prolongs design iteration cycles and severely impedes early-stage Design-Technology Co-Optimization (DTCO), as shown in \Cref{fig:workflow}.

Recent advances in machine learning \cite{jiang2024fabsage,jiang2024fabgpt,jiang2025fabthink,jin2024semclip,qiao2024minimizing,zhou2019efficient,pan2023lithography} offer promising solutions to this precision-speed tradeoff. Existing research, often offered as a computer vision task, leverages data-driven models to approximate the outcomes of traditional simulations \cite{huang2021machine}. They capitalize on the pattern recognition strengths of neural networks to identify critical features that traditional rule-based methods might overlook. In lithography simulation, researchers have utilized generative models like generative adversarial networks (GANs) for mask and contour generation \cite{yang2018gan,ye2019lithogan,chen2020damo,shao202LithoNet,zhang2023litho}. Similarly, in hotspot detection, deep learning has been employed to pinpoint defect locations 
\cite{chen2022faster, wu2024enhancing,shao2024lithohod}. These ML-based methods can reduce simulation times from hours to minutes or even seconds for specific tasks, enabling faster iterations during design validation.

Despite these advancements, significant limitations persist in these existing approaches. First, current methods exhibit a single-task orientation. Many focus exclusively on mask optimization, lithography simulation, or hotspot detection, neglecting information flow across design, simulation, and verification stages. This fragmentation forces designers to rely on disparate tools and workflows, complicating the design process and introducing inefficiencies. Second, these models suffer from limited adaptability to multi-process conditions. Typically trained on fixed process conditions, they struggle to generalize across the diverse conditions encountered in real-world fabrication. This limitation also hampers their applicability to process window analysis, which evaluates design robustness across a range of conditions, and PVBand simulation, which assesses contour variability under process fluctuations. 

To address these critical challenges, we propose Unitho, a novel, integrated framework that unifies lithography simulation and detection. It is designed as an automated tool to rapidly assess the manufacturing feasibility of a design across diverse process conditions. Additionally, it incorporates a real-time detector to identify and highlight DRC, MRC, and LRC violations with highlighted bounding boxes. Our method can perform a full process window simulation in minutes, providing designers with ``Lithography-Friendly-Design (LFD) Clean'' feedback before tape-out, significantly shortening the find-to-fix loop. The contributions are summarized as follows:

\begin{itemize}
  \item We propose Unitho, a unified multi-task large vision model designed for computational lithography optimization, integrating end-to-end mask generation, lithography simulation, and violations detection. 
  \item We fuse process conditions as contextual embeddings with layout features through a cross-attention fusion module, enabling rapid and precise generation of masks and contours across diverse process conditions.
  \item We design a contrastive learning strategy for the generation that enhances structural distinctiveness across layouts and edge precision under diverse conditions, improving process sensitivity and generalization.
  \item Leveraging multi-scale feature fusion and a customized loss function tailored for hotspot detection tasks, Unitho delivers immediate, accurate, and visualized feedback on potential violations.
  \item We evaluate the proposed techniques on an in-house industrial dataset, containing 624{,}129 layout clips and the corresponding masks and contours. The results show that our method is $85\times$ faster than the commercial tool.
\end{itemize}

  


%% file: figs/runtime.tex
\pgfplotsset{
    width=0.9\linewidth,
    height=0.38\linewidth
}
\begin{tikzpicture}
    \begin{axis}[
        xbar stacked, 
        xmode=log,
        xmin=10, xmax=15000,
        bar width=0.35cm,
        yticklabels={Ours (104.96 ms), Commercial Tool (8913.1 ms)},
        yticklabel style={align=center, text width=1.4cm, font=\footnotesize},
        ytick={1, 3},
        ymax=4, ymin=0,
        xtick={10, 100, 1000, 10000},
        xticklabels={10, 100, 1000, 10000},
        xticklabel style={font=\small},
        xlabel={Runtime (ms, log$_{10}$ scale)},
        xlabel style={font=\small},
        legend style={draw=none, at={(0.45,1.02)}, anchor=south, legend columns=3, font=\small},
        grid=major,
        grid style={dashed, gray!30}
    ]
    \addplot [fill=darkblue] coordinates {(49.25,1) (3520.34,3)};
    \addplot [fill=lightblue] coordinates {(49.25,1) (4211.56,3)};
    \addplot [fill=CUribbon] coordinates {(6.46,1) (1181.24,3)};
    
    \node[font=\small, right] at (axis cs:350,1) {\textbf{×85 Faster}};
    
    \legend{OPC\,\,\,,\,Simulation\,\,\,,\,Verifications}
\end{axis}
\end{tikzpicture}

%% file: doc/prelim.tex
\section{Preliminaries}
\label{sec:preliminaries}

\subsection{Post-Layout Manufacturability Verification}

Physical verification ensures that layouts and masks comply with manufacturing specifications, reducing yield loss and reliability issues. It includes three complementary checks: DRC, MRC, and LRC.

\textbf{DRC}
verifies whether the layout complies with process-defined geometric rules, such as minimum width, spacing, and area, derived from physical fabrication limits, such as lithographic resolution constraining minimum widths. Applied post-placement and routing in the physical design phase, DRC serves as the foundational step in verification, identifying violations early to prevent downstream defects.

\textbf{MRC} is conducted following mask synthesis techniques, such as OPC, to verify the manufacturability of photomasks. It enforces constraints including minimum feature sizes, spacing ratios, and shape restrictions,
thereby ensuring adherence to mask manufacturing rules and optimizing the yield and overall performance of the integrated circuit (IC).

\textbf{LRC}, often referred to as optical proximity correction verification, evaluates the layout's printability under lithographic conditions. It detects hotspots—regions where printed patterns deviate from the intended design—and variations in critical dimension (CD) via rigorous simulations. Conducted in the post-OPC 
validation phase, LRC ensures wafer-level fidelity by modeling optical effects and process variations.


\subsection{Process Window in Computational Lithography}

The process window defines the multidimensional range of key manufacturing parameters, primarily exposure dose and focus, within which a lithographic pattern can be printed on the wafer while meeting specified tolerances, such as CD uniformity and edge placement error (EPE) tolerances. This metric quantifies the robustness of IC designs against process variations, with a larger PW indicating higher yield potential by accommodating real-world fluctuations in fabrication conditions.

PW plays an essential role in LFD workflows, which aim to enhance 
manufacturability through process window verification (PWV) by evaluating pattern printability under varied process conditions. Specifically, tools simulate resist contours at multiple PW points, checking compliance with design rules. If non-compliant, such as 
bridging or pinching, the analysis triggers feedback loops for layout modifications, such as adjusting feature geometries 
during OPC. This integration with design for manufacturing (DFM) principles enables predictive yield estimation and iterative refinement, preventing costly downstream failures. Nonetheless, conventional PW analysis often involves computationally expensive simulations, which can take hours to days per iteration, thereby limiting scalability for advanced nodes.

\subsection{Lithography Simulation}
\label{sec:Lithography Simulation}

Conventional lithography simulation employs physics-based models, dividing the process into optical imaging and photoresist modeling \cite{Jin25CLT}.
Recent machine learning methods accelerate this via image-to-image translation, often using simplified one-way mappings (e.g., mask-to-contour or layout-contour). For instance, LithoGAN \cite{ye2019lithogan} uses cGAN \cite{isola2017image} for direct contour prediction from masks, bypassing explicit physics. DAMO \cite{chen2020damo} employs DCGAN-HD for high-resolution contour generation with multi-scale discriminators. LithoNet \cite{shao202LithoNet} combines CycleGAN for transformation and U-Net for deformation learning, incorporating process conditions for adaptability. 
However, these methods ignore couplings between layout, mask, and process parameters, resulting in incomplete pipelines, limited controllability under variations, and poor applicability in cross-process or feedback-driven scenarios.

\subsection{Hotspot Detection}
\label{sec:Hotspot Detection}

Deep learning has accelerated hotspot detection in lithography by framing it as classification or object detection. Early methods, with advancements such as attention-driven CNNs, have improved accuracy through joint training \cite{geng2020hotspot}. Sun et al. \cite{sun2022efficient} used graph neural networks (GNNs) for geometric embeddings, Lin et al. \cite{lin2022lithography,pan2023lithography} used heterogeneous federated learning, combining global and local submodels with local adaptation, 
and Xu et al. \cite{xu2024lithography} integrated attention modules and feature pyramids for multi-scale fusion.

More recent studies have shifted from simple hotspot classification to precise localization. Two-stage frameworks like Faster R-CNN, adapted by Chen et al. \cite{chen2022faster} are used for bounding-box refinement. Single-stage detectors like augmented YOLOv8 leverage PCA-guided augmentation to enhance accuracy \cite{wu2024enhancing}. To further improve generalization, research integrates physical simulation with deep learning. Shao et al. \cite{shao2024lithohod} fuse lithographic simulator outputs with a single-stage object detection network via cross-attention mechanisms. Despite these advancements, existing methods remain confined to single-process condition scenarios and focus mainly on basic LRC classification, lacking generalization across conditions and comprehensive violation detection in early design.

\subsection{Problem Definition}
In this work, we formalize AI-assisted computational lithography as three complementary tasks: mask/contour generation, violation detection, and integrated generation-detection for rapid PWV. To unify terminology, we refer to regions failing DRC, MRC, or LRC as ``violations'' or ``hotspots''.

\begin{myproblem} [Generation]
Given an input layout clip $L$ and associated process conditions $P$ (including the source image, dose, focus, and photoresist threshold), the task is to generate the corresponding mask $\hat{G_M}$ and resist contour $\hat{G_C}$ in a single forward pass. 
The objective is to minimize discrepancies with the ground-truth mask $G_M$ and contour $G_C$, evaluated using metrics that assess pixel-level accuracy, regional overlap, and boundary fidelity, as detailed in \Cref{sec:Evaluations}.
\end{myproblem}

\begin{myproblem} [Detection]
Given the input layout clip $L$ for DRC hotspot detection, a mask clip $G_M$ for MRC hotspot detection, or the pair of a contour clip $G_C$ and $L$ for LRC hotspot detection, identify and localize violations using bounding boxes. 
The goal is to ensure precise detection for iterative design refinement, measured by detection reliability and error minimization metrics, detailed in \Cref{sec:Evaluations}.
\end{myproblem}

\begin{myproblem} [Unified Detection]
To enable rapid process-aware LRC hotspot prediction, we integrate generation and detection 
by fine-tuning the detector on outputs from the generative module. Specifically, given an input layout clip $L$ and process conditions $P$, we can directly detect and localize LRC hotspots under varying conditions, providing immediate feedback on lithography-related violations without relying on simulations. 
\end{myproblem}

%% file: doc/algo.tex
\section{Proposed Method}
\label{sec: method}

\begin{figure*}[t]
    \centering
    \includegraphics[width=\linewidth]{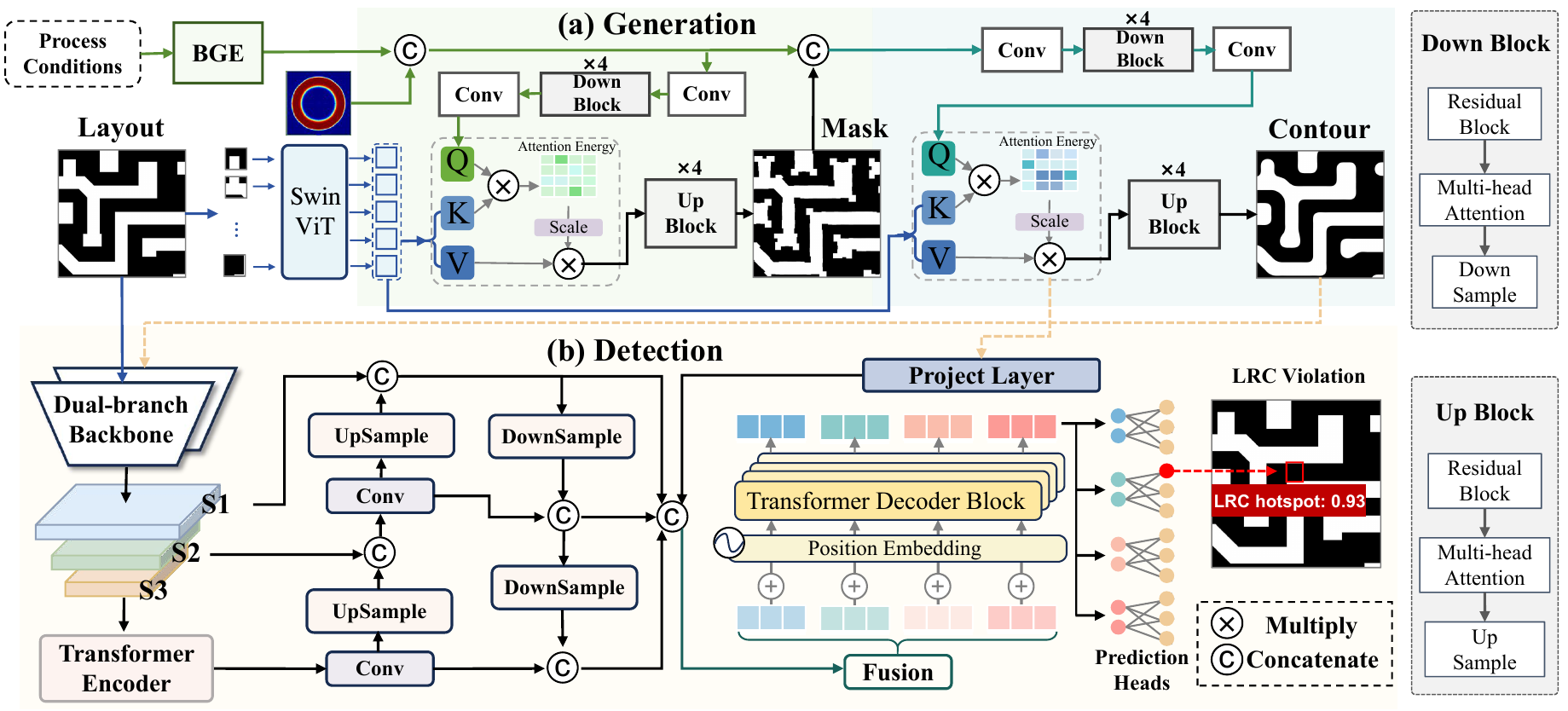}
    \caption{The framework of Unitho. }
    \label{fig:framework}
\end{figure*}

As shown in \Cref{fig:framework}, the Unitho framework integrates a generative model and a detection model for efficient process-aware LRC hotspot prediction. The generative model uses a cross-attention fusion and contrastive learning to produce high-fidelity masks and contours across multiple process conditions. In parallel, the detection model incorporates multi-scale feature fusion,
a single-layer transformer for computational efficiency, and a tailored loss function to reduce false positives and improve recall. By blending generative semantic features into the detection pipeline and conducting joint fine-tuning, the Unitho framework achieves end-to-end prediction, with fast and accurate localization of potential violations 
across process conditions.

\subsection{Unified Lithography Simulator}
We present a process-aware end-to-end generator that converts the layout under different process conditions
directly into the mask and contour in a single inference step.
This architecture consists of four core components: a layout encoder, a process parameter encoder, a cross-attention fusion module, and an image decoder, as detailed below.

\minisection{Layout Encoder}
The geometric patterns in the layout determine the printed images. To capture long-range dependencies and fine structures, we employ a Swin Transformer encoder. Unlike conventional CNNs, its hierarchical window-based attention extracts local details and global context simultaneously.

The layout image is first partitioned into fixed-size patches, which are flattened and linearly projected into token vectors augmented with positional embeddings to retain spatial structure. These tokens traverse the multi-stage Swin-Transformer encoder: windowed self-attention captures local semantics, the shifted-window scheme enables cross-region context, and the hierarchical stages fuse multi-scale features. The resulting layout embedding forms the structural basis for subsequent mask and contour generation.

\minisection{Process Condition Encoder}
To enhance adaptability to variations in real-world lithography conditions, we design a process condition encoding module. This module takes source images and three scalar parameters
as inputs. 
Scalar parameters are embedded into feature maps matching the source image resolution along the channel dimension using a lightweight BGE model \cite{xiao2024c}. These maps are concatenated with the source image to form a four-channel tensor, then processed by an encoder comprising alternating convolutional and attention layers to capture semantic interactions among these process conditions.

For contour generation, the predicted mask is appended as a fifth channel, and the resulting tensor is encoded by the same network to extract updated process-aware features. Joint reasoning over the mask and process conditions improves the physical consistency and accuracy of contour prediction.

\minisection{Cross-Attention Fusion Module}
\label{sec:fusion}
To establish semantic linkage between layout structures and process-aware features, we introduce a multi-head cross-attention mechanism to capture how layout patterns respond under varying process conditions. The core idea of this module is to guide the model to selectively focus on layout regions that are most sensitive to imaging outcomes, thereby enhancing adaptability and interpretability under different lithography settings. Specifically, embeddings from the process parameter encoder serve as the query $Q_P$, while the layout structural embeddings from the layout encoder act as the key $K_L$ and value $V_L$. The cross-modal attention is computed as:
\begin{align}
\text{CrossAttn}(Q_P, K_L, V_L) = \text{Softmax} \left( \frac{Q_P K_L^\top}{\sqrt{d_k}} \right) V_L,
\end{align}
where $d_k$ denotes the dimensionality of the key vectors. 

\minisection{Image Decoder}
The image decoder reconstructs the fused semantic features into their corresponding mask or contour images.
We adopt a decoder architecture integrating transposed convolutional layers, residual blocks, and multi-head attention units. This combination preserves local details while incorporating global semantics, enhancing the quality of pattern reconstruction.

The decoder first generates the mask from the fused features. The mask is then appended to the process features, and the decoder runs once more to produce the contour image. This dependent generation occurs within a single inference pass, maintaining computational efficiency.

The proposed generation model emphasizes structure-process co-modeling and explicitly maps conditional dependencies among layout, mask, and contour. By employing a cross-attention mechanism to couple layout structures with process conditions and leveraging intermediate feature guidance for process-aware generation, the model achieves robust generalization and physical consistency. 

\subsection{Fast Hotspots Detector}
Current CNN-based hotspot detection methods rely on local receptive fields, limiting their ability to capture global and long-range dependencies essential for geometric pattern modeling. Additionally, they generate redundant bounding boxes, necessitating computationally costly Non-Maximum Suppression (NMS). To address this, we utilize Transformers' attention mechanisms for global context modeling and a modified RT-DETR \cite{zhao2024detrs} architecture for efficient, high-speed detection.

The feature encoding process commences with inputs tailored to each task. These inputs are processed by a lightweight ResNet-18 backbone, which leverages residual connections to extract hierarchical features, yielding multi-scale feature maps labeled $S1$ through $S3$. These maps capture varying levels of abstraction: high-resolution maps like $S1$ provide fine-grained details, such as precise edges and minute geometric anomalies critical for detecting localized violations, like EPE errors; conversely, low-resolution maps like $S3$ encode global semantics essential for contextualizing LRC violations influenced by neighboring patterns and optical proximity effects. To optimize computational efficiency, we employ a single-layer transformer encoder for $S3$, where the flattened feature maps are refined through self-attention to capture long-range dependencies across the input. Moreover, to preserve hierarchical information, we employ upsampling and downsampling operations: higher-level features $S1$ are downsampled to propagate global context to finer scales $S3$, while lower-level features $S3$ are upsampled to enrich global representations with local details $S1$, and the resulting feature maps are fused via concatenation to create a comprehensive input for the decoder. The decoder follows the standard DETR paradigm \cite{carion2020end}, employing learnable object queries interacting with the encoded features via cross-attention, iteratively refining their focus on potential violations across multiple decoder layers. A prediction head maps each query to bounding box coordinates and classification labels (violations or no-hotspot), directly generating detections without NMS, thereby enhancing inference speed. 


For DRC and MRC, the Detector processes a single input. In contrast, LRC requires assessing deviations between the resist contour and the target layout, necessitating joint processing of both inputs. To accommodate this, we modify the backbone into a dual-branch structure: separate initial features are extracted from contour and layout, which are then concatenated and fused to produce a unified feature map. This fused representation proceeds through the shared residual layers, preserving computational efficiency. This adaptation aligns with the semantic characteristics of LRC violations and enhances detection robustness early in feature extraction.


\subsection{Unified Generation-Detection Framework}

Building on the above, we propose a joint generation-detection framework for end-to-end process-aware LRC hotspot prediction. This integration leverages the generative model's outputs as inputs for the detection model, enabling rapid PWV without relying on time-consuming physical simulations. Specifically, the generative model produces the predicted contour $\hat{G_C}$, along with the corresponding layout, which is then input into the LRC-specific dual-branch detector for hotspot localization, yielding bounding boxes around potential LRC violations.

To further enhance detection accuracy in this integrated setting, we incorporate semantic features from the generative model into the detection pipeline. During generation, the cross-attention fusion module aggregates process-conditioned layout embeddings, yielding high-level semantic features that capture optically relevant patterns and proximity effects, as detailed in \Cref{sec:fusion}. These features are projected through a lightweight projection layer and then concatenated with the encoder feature, followed by a fusion convolution that integrates generative insights directly into the detection backbone. This cross-module feature fusion bridges the generative and detection tasks, providing the detector with contextual priors from the simulation process, which improves hotspot localization under generated contours and enhances generalization to diverse process conditions.

\subsection{Training strategy}
Our training consists of three phases, aiming to build an integrated framework for computational lithography tasks. First, we pre-train the generative model to accurately produce masks and contours from layout clips and process conditions. Second, we pre-train the detector independently to detect hotspots with high precision. Finally, to realize end-to-end PWV assessment, we conduct fine-tuning that combines contour generation with LRC hotspot detection, adapting the detector to the generated outputs.

\subsubsection{Generation Pre-training}
To enable high-quality generation of both masks and contours while accurately capturing variations in lithography process conditions, we adopt a joint training strategy that simultaneously optimizes these two generation tasks. We propose a loss function that consists of two complementary components: a reconstruction loss and a contrastive learning objective. This formulation ensures both fidelity of the generated outputs and sensitivity to variations in lithography process conditions. 

The total loss is defined as a weighted combination of the two objectives:
\begin{align}
\mathcal{L}_{\text{total}}^G = 
w_{\text{rec}} \cdot 
\mathcal{L}_{\text{reconstruction}} + w_{\text{con}} \cdot \mathcal{L}_{\text{contrast}},
\end{align}
where $\mathcal{L}_{\text{reconstruction}}$ encourages high-fidelity generation of masks and contours, and $\mathcal{L}_{\text{contrastive}}$ promotes the model’s ability to discriminate between different process conditions and input layouts.

\minisection{Reconstruction Loss}
The reconstruction loss is designed to improve the consistency between the generated outputs and their corresponding ground truth, comprising two components: the mask generation loss and the contour generation loss. Specifically, it is defined as
\begin{align}
\mathcal{L}_{\text{reconstruction}} = w_{\text{mask}} \cdot  \mathcal{L}_{\text{mask}} + w_{\text{contour}} \cdot  \mathcal{L}_{\text{contour}}.
\end{align}

Given that contour prediction depends on the intermediate outputs of mask generation, we allocate a higher weight to the mask loss term to prioritize the production of high-fidelity masks, thereby enhancing the accuracy of subsequent contour predictions. And the definitions of $\mathcal{L}_{\text{mask}}$ and $\mathcal{L}_{\text{contour}}$ are similar; both are composed of a weighted combination of three components:
\begin{multline}
\mathcal{L}_{\text{mask}} = 
w_{\text{Dice}} \cdot \mathcal{L}_{\text{Dice}} + 
w_{\text{BCE}} \cdot \mathcal{L}_{\text{BCE}} + 
w_{\text{Edge}} \cdot \mathcal{L}_{\text{Edge}},
\end{multline}
and $\mathcal{L}_{\text{contour}}$ follows this formula.

Let $G$ denote the ground truth binary mask or contour, and $\hat{G}$ the corresponding prediction. The Binary Cross Entropy (BCE) loss optimizes pixel-wise accuracy:
\begin{align}
\mathcal{L}_{\text{BCE}} = - \big[ G \cdot \log(\hat{G}) + (1 - G) \cdot \log(1 - \hat{G}) \big].
\end{align}

The Dice loss focuses on region-level overlap and structural similarity:
\begin{align}
\mathcal{L}_{\text{Dice}} = 1 - \frac{2 \cdot |\hat{G} \cap G|}{|\hat{G}| + |G|}.
\end{align}

To further enhance boundary sensitivity, we introduce an edge-aware Dice loss. Specifically, edge regions are extracted from $G$ and $\hat{G}$ via morphological erosion, denoted as $\text{Edge}(\cdot)$, and the Dice similarity is computed within these regions:
\begin{align}
\mathcal{L}_{\text{Edge}} = 1 - \frac{2 \cdot |\text{Edge}(\hat{G}) \cap \text{Edge}(G)|}{|\text{Edge}(\hat{G})| + |\text{Edge}(G)|}.
\end{align}

These losses are computed on positive samples (matched layout-condition pairs) to focus on accurate generation.

\minisection{Contrastive Learning Objective}
For the same layout, predictions under matched process conditions should closely resemble the corresponding ground truth to ensure accuracy. In contrast, under mismatched conditions, the generated image should exhibit distinguishable structural differences that reflect the specific characteristics 
To achieve this and enhance the model’s sensitivity to process conditions, we incorporate two complementary contrastive objectives: Structure-Aware Contrast (SAC) and Process-Aware Contrast (PAC). The total contrastive loss is defined as:
\begin{align}
\mathcal{L}_{\text{contrast}} = \mathcal{L}_{\text{SAC}} + \mathcal{L}_{\text{PAC}}.
\end{align}

SAC promotes global discrimination by strongly pushing apart predictions from different layouts, ensuring that representations remain distinct across unrelated inputs. For a given layout $L$ under process $P$, the anchor is the prediction $\hat{G}(L, P)$ from a matched pair, while negatives include predictions from other layouts $\tilde{L}$ under any process $\hat{P}$, formulated as:
\begin{align}
\mathcal{L}_{\text{SAC}} = 
& - \frac{1}{|\mathcal{P}|} \sum_{i \in \mathcal{P}} 
& \log \frac{ 
\exp \left( \frac{\text{Dice} \left( \hat{G}_i(L, P), \hat{G}_i(L, \hat{P}) \right)}{\tau} \right)
}{
\sum\limits_{j \in \mathcal{N}} 
\exp \left( \frac{\text{Dice} \left( \hat{G}_i(L, P), \hat{G}_j(\tilde{L}, \hat{P}) \right)}{\tau} \right)
},
\end{align}
where $\mathcal{P}$ denotes the set of anchors from matched pairs, $\mathcal{N}$ denotes negatives from different layouts, and $\tau = 0.07$ is the temperature parameter. This encourages strong dissimilarity between outputs of distinct layouts.

To handle fine-grained distinctions for the same layout under different process conditions, where outputs are globally similar but should preserve boundary-level variations, we introduce PAC. It penalizes excessive similarity in boundary regions, encouraging process-sensitive differences:
\begin{multline}
\mathcal{L}_{\text{PAC}} = \frac{1}{|\mathcal{K}|} \sum_{i \in \mathcal{K}} \text{ReLU}\!\left( m - \mathcal{L}_{\text{Edge}}\!\left( \hat{G}_i(L,\!\hat{P}), G_i(L,\!P) \right) \right),
\end{multline}
where $\mathcal{K}$ denotes the set of prediction pairs from the same layout under varying process conditions, used to compute edge-based similarity with the ground truth. $\mathcal{L}_{\text{edge}}(\cdot)$ measures Dice similarity on edge regions and $m$ is a margin threshold set to 0.5. 
This term helps the model retain process-dependent geometric nuances while maintaining overall structural consistency.

\subsubsection{Detection Pre-training}
In the pre-training phase of the detector, to address challenges of class imbalance, defect sparsity, and high localization precision requirements in hotspot detection tasks, we introduce the detection loss, formulated as follows:
\begin{multline}
\mathcal{L}_{\text{total}}^D = w_{\text{vfl}} \cdot \mathcal{L}_{\text{vfl}} + w_{\text{fppl}} \cdot \mathcal{L}_{\text{fppl}} + w_{\text{bbox}} \cdot \mathcal{L}_{\text{bbox}} + w_{\text{giou}} \cdot \mathcal{L}_{\text{giou}}.
\label{eq:detector}
\end{multline}


\minisection{Varifocal Loss}
Severe class imbalance in hotspot detection biases models toward negatives, causing missed violations or high false alarms. To mitigate this, we adopt Varifocal Loss \cite{zhang2021varifocalnet}, a variant of BCE, for hotspot classification.  VFL weights samples by prediction quality, using IoU-based targets for positives to enhance precision and suppress low-quality negatives. It is defined as:
\begin{align}
\mathcal{L}_{\text{vfl}} = \text{BCE}(p, t, w) = -w \left[ t \log(p) + (1-t) \log(1-p) \right],
\end{align}
where $p$ is the predicted score, $t$ the target score ($t = \text{IoU} \cdot \text{one-hot label}$), and $w = \alpha p^\gamma (1-t) + t$ ($\alpha = 0.75$, $\gamma = 0.5$). 

\minisection{False Positive Penalty Loss}
In violation hotspot detection, false positives can lead to unnecessary manufacturing adjustments, escalating costs, and delaying production cycles. To minimize these, we introduce False Positive Penalty Loss (FPPL), which specifically penalizes high-confidence non-hotspot regions, thereby encouraging the reduction of false positives and optimizing the FA-recall trade-off.

Inspired by Focal Loss variants for negative samples, FPPL is applied exclusively to negatives and formulated as:
\begin{align}
\mathcal{L}_{\text{fppl}} = -\alpha (1 - p)^\gamma \log(1 - p),
\end{align}
here $p$ is the predicted score for negative samples, $\alpha = 0.25$, $\gamma = 2.0$. High-confidence false positives are penalized more.

\minisection{Regression Loss}
Another key challenge lies in precisely localizing hotspot regions. In our task, hotspots may exhibit irregular shapes or partial overlaps. Standard regression losses suffer from insufficient gradients in non-overlapping cases. To address this, we combine L1 loss with Generalized IoU (GIoU) loss for bounding box regression. The L1 loss provides uniform coordinate optimization, formulated as:

\begin{align}
\mathcal{L}_{\text{bbox}} = \frac{1}{n} \sum_{i=1}^{n} \|\hat{b}_i - b_i\|_1 ,
\end{align}
where $\hat{b}_i$ and $b_i$ represent the predicted and ground-truth bounding boxes for the $i$-th sample, $n$ is the total number of samples.

GIoU adds geometric constraints, is defined as:
\begin{align}
\mathcal{L}_{\text{giou}} = \left[1 - \left( \mathrm{IoU}(\hat{b}_i, b_i) - \frac{|R_i \setminus (\hat{b}_i \cup b_i)|}{|R_i|} \right)\right],
\end{align}
with $R_i$ denotes the smallest enclosing rectangle for the $i$-th pair. This enhances accuracy for sparse, complex violations. 

\subsubsection{Joint Fine-tuning}
In this phase, we freeze the weights of the pre-trained generative model to maintain its predictive stability, employing its generated contours as inputs to the detection model. This step enables the detection model to adapt to the characteristics of the generative model's outputs, thereby optimizing its performance in localizing LRC hotspots within an end-to-end framework. The fine-tuning process utilizes the same loss function established in the detection pre-training phase, as defined in Equation \ref{eq:detector}, ensuring continuity in the optimization strategy across training stages.

%% file: doc/result.tex
\section{Experiment}
\label{sec:exp}
\subsection{Experimental Setup}
\minisection{Datasets}
We construct a large-scale lithography dataset at the $55\,nm$ technology node using commercial tools. Each image is $512 \times 512$ pixels, representing $36\,\mu\mathrm{m}^2$ of physical area.

To improve the model’s robustness against process variations, we consider
four key process conditions: source type (Annular, Circular, and Bull’s Eye), resist threshold (three representative photoresists), focus ($0\,nm$, $50\,nm$), and exposure dose ($1.0\times$ and $1.2\times$). These conditions form $36$ process combinations, yielding $624{,}129$ layout–mask–contour triplets.


Furthermore, we use commercial tools to label lithographic hotspots at three stages—$63{,}558$ layout images with DRC violations, $55{,}690$ mask images with MRC violations, $116{,}881$ contour images with LRC violations, and an additional $100{,}000$ no-hotspot triplets. 
These triplets are used for generative modelling and detection tasks.  
For detection, we follow the train/test splits in \Cref{tab:dataset}; the $100{,}000$ no-hotspot triplets are shared across tasks.


\begin{table}[b]
    \centering
    \caption{Dataset split for DRC, MRC, and LRC tasks.}
    \resizebox{0.94\linewidth}{!}
    {
        \begin{threeparttable}
            {
                \begin{tabular}{lcccc}
                    \toprule
                    Task & {\centering Train \#Clips} & {\centering Train \#HS} & {\centering Test \#Clips} & {\centering Test \#HS} \\
                    \midrule
                    DRC & 140558 & 44271  & 23000 & 8460 \\
                    MRC & 129890 & 74642  & 25800 & 15847 \\
                    LRC & 180281 & 196845 & 36600 & 42346 \\
                    \bottomrule
                \end{tabular}
            }
        \end{threeparttable}
    }
    \label{tab:dataset}
\end{table}

\minisection{Implementation Details}
For the pre-training of the generation task, we configured the loss weights as follows: $w_{\text{rec}} = 1$, $w_{\text{con}} = 0.5$, $w_{\text{mask}} = 2$, $w_{\text{contour}} = 1$, $w_{\text{Dice}} = 2$, $w_{\text{BCE}} = 1$, and $w_{\text{Edge}} = 0.3$. For the pre-training of the violation detection task, the loss weights were set to: $w_{\text{vfl}} = 1$, $w_{\text{bbox}} = 5$, $w_{\text{giou}} = 2$, and $w_{\text{fppl}} = 1.5$. All experiments were conducted on $8$ NVIDIA A100 GPUs.

\minisection{Evaluations}
\label{sec:Evaluations}
For the generation task, we focus on pixel-level and Edge-specific measures to capture both global fidelity and edge precision. Here, $\hat{y}_i$ and $y_i$ are the predicted and ground-truth labels for pixel $i$ in $\hat{G}$ and $G$.

\textbf{Mean Pixel Accuracy (mPA)} measures the overall correctness of pixel-level predictions across the entire image:
\begin{align}
 \text{PA} = \frac{\sum_{i=1}^{M} \mathbb{I}(y_i = \hat{y}_i)}{M}, \quad
 \text{mPA} = \frac{1}{N} \sum_{j=1}^{N} \text{PA}_j,
\end{align}
where $M$ is the total number of pixels, $N$ is the number of test examples, $\mathbb{I}$ is the indicator function, which takes the value of 1 when $y_i = \hat{y}_i$ and 0 otherwise. 

\textbf{Mean Intersection over Union (mIoU)} quantifies average region overlap:  
\begin{align}
  \text{mIoU} = \frac{1}{N}\sum_{i=1}^N \frac{|G_i \cap \hat{G}_i|}{|G_i \cup \hat{G}_i|}.
\end{align}

\textbf{Edge $F_1$-Score} evaluates contour/mask edge precision using morphological dilation, $\text{Edge F}_{1} = \frac{2 \times \text{P} \times \text{R}}{\text{P} + \text{R}}$:
\begin{align}
  \text{P} = \frac{|\text{Edge}(\hat{G}) \cap \text{Edge}(G)|}{|\text{Edge}(\hat{G})|}, \quad
  \text{R} = \frac{|\text{Edge}(G) \cap \text{Edge}(\hat{G})|}{|\text{Edge}(G)|}.
\end{align}

In our detection task, true positive (TP) is defined as the number of predicted bounding boxes that correctly match ground-truth violations, based on IoU overlap exceeding 0.5 and confidence above 0.6; false positive (FP) represents predicted boxes that do not match any ground-truth hotspot, corresponding to erroneous detections of non-violation regions. \textbf{Recall} evaluates the completeness of hotspot detection. \textbf{Precision} indicates detection reliability. And \textbf{$F_1$-score} is the harmonic mean of them. \textbf{\#FA} equals FP, quantifying erroneous detections. \textbf{Average Precision at IoU=0.5 (AP@0.5)} is area under precision-recall curve with IoU threshold 0.5.

\subsection{Experimental Results}
\label{sec:result}
\begin{figure}[tb]
    \centering
    \includegraphics[width=\linewidth]{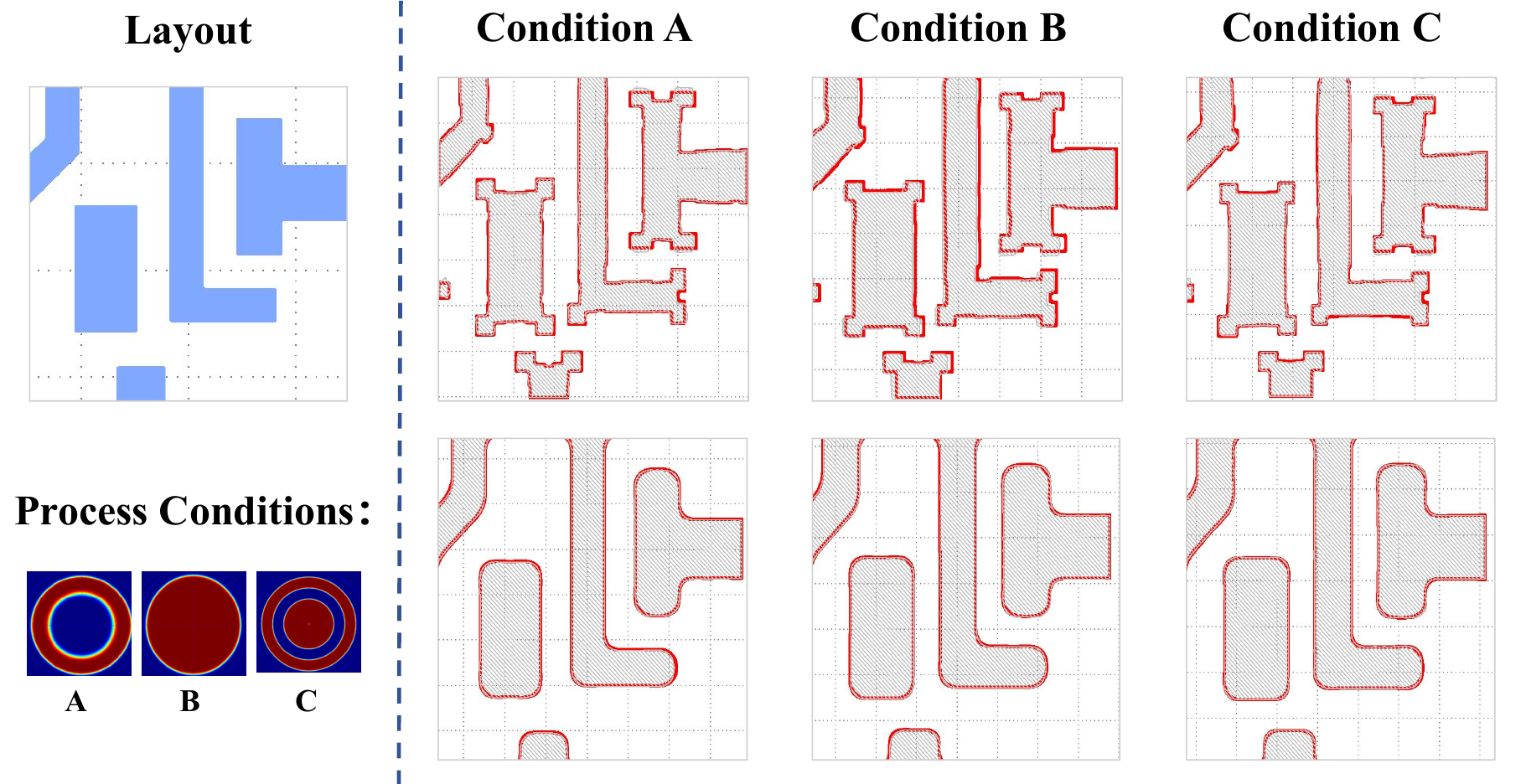}
    \caption{
    Generated results under representative process conditions, with each pair showing predicted mask (top) and contour (bottom).
}
    \label{fig:Generate}
\end{figure}

\minisection{Generation Results}
We apply the Unitho framework to the end-to-end generation of mask and contour images under 36 distinct process conditions. \Cref{fig:Generate} shows the generation results under three representative process conditions, where A, B, and C correspond to the first three settings listed in \Cref{tab:LRC}. For example, ``Annular\_0.0923125\_50\_1'' denotes an Annular source type with resist threshold of 0.0923125, focus of 50nm, and dose of 1.0$\times$.
The red outlines represent outputs from commercial tools, while Unitho generates the gray regions. As seen, Unitho generates masks and contours that closely match the labeled shapes, and it is capable of producing different edge details for the same layout under varying process conditions.

We further benchmark Unitho against several representative lithography modeling methods, detailed in \Cref{sec:Lithography Simulation}. All methods are evaluated under consistent data splits and training protocols to ensure fairness and comparability.
We evaluate performance using evaluations defined in \Cref{sec:Evaluations}.
 As shown in \Cref{tab:Comparison-of-generation}, Unitho achieves the best performance across all metrics, significantly exceeding prior methods.

\begin{table}[tb]
  \centering
  \caption{Comparison of evaluation metrics among different lithography modeling methods on generation task (\%).}
    {
        \begin{threeparttable}
        \centering
\tabcolsep=3pt
\renewcommand{\arraystretch}{1.1}
        {
\resizebox{0.92\linewidth}{!}{%
\begin{tabular}{c|ccc|ccc}
\hline
\multirow{2}{*}{Methods }  & \multicolumn{3}{c|}{Mask Generation}  & \multicolumn{3}{c}{Contour Generation}   \\ \cline{2-7} 
 &
  \multirow{1}{*}{mPA ↑} & 
  \multirow{1}{*}{mIoU ↑} &
  \multirow{1}{*}{$F_1$ ↑} & 
  \multirow{1}{*}{mPA ↑} &
  \multirow{1}{*}{mIoU ↑} &
  \multirow{1}{*}{$F_1$ ↑}   \\ \hline
CGAN~\cite{isola2017image} & 97.80 & 94.02 & 97.10  & 98.06 & 95.15 & 99.21 \\
LithoNet~\cite{shao202LithoNet}   & 96.06 & 89.65 &  91.38  & 91.10 & 78.31 & 53.83  \\  
\textbf{Unitho}    & \textbf{98.96}  &  \textbf{96.98} & \textbf{99.08}  & \textbf{99.25}  & \textbf{99.23}   &  \textbf{99.76}         \\ \hline
\end{tabular}
}
}
        \end{threeparttable}
    }
  \label{tab:Comparison-of-generation}
\end{table}

To ensure that the generated contour introduces no new LRC violations, we converted the contour images produced by Unitho into GDS files and submitted them to a commercial tool for LRC verification. \Cref{tab:LRC} presents the LRC violation counts across five distinct process conditions,  compared against the results obtained from the original simulation tool using the same layout inputs. 
Our analysis reveals that the defect counts from Unitho’s contours closely match those of the standard simulation tool, with deviations consistently remaining below $4\%$. This highlights Unitho’s ability to maintain high-fidelity lithographic responses and capture fine-grained geometric variations introduced by process changes, ensuring its feasibility for downstream detection tasks.

\begin{table}[tb]
  \centering
  \caption{Comparison of LRC violations for generated and simulated contours.}
    {
        \begin{threeparttable}
        \centering
\tabcolsep=3pt
\renewcommand{\arraystretch}{1.1}
        {
\resizebox{\linewidth}{!}{%
\begin{tabular}{c|ccc|ccc}
\hline
\multirow{2}{*}{\shortstack{{Process} \\ {Conditions}}}
 & \multicolumn{3}{c|}{The Commercial Tool}  & \multicolumn{3}{c}{Unitho (ours)}   \\ \cline{2-7} 
 &
  \multirow{1}{*}{\#Pinch} & 
  \multirow{1}{*}{\#Bridge} &
  \multirow{1}{*}{\#EPE} &
  \multirow{1}{*}{\#Pinch} & 
  \multirow{1}{*}{\#Bridge} &
  \multirow{1}{*}{\#EPE}  \\ \hline
A\tnote{+}  & 1182 & 2898 &  5902 & 1180(-0.2\%) & 2926(+1.0\%) &  6010(+1.8\%)  \\    
B\tnote{+}  & 1326 & 3099 &  6322 & 1302(-1.8\%) & 3118(+0.6\%) &  6551(+3.6\%)  \\
C\tnote{+}  & 1228 & 3284 & 14228 & 1212(-1.3\%) & 3237(-1.4\%) & 14190(-0.2\%)  \\
D\tnote{+}  & 1247 & 2902 &  6099  & 1232(-1.2\%) & 2917(+0.5\%) &  6338(+3.9\%)  \\
E\tnote{+} & 1129 & 3206 & 14202 & 1116(-1.2\%) & 3160(-1.4\%) & 14148(-0.3\%)  \\ \hline
\end{tabular}
}
            \begin{tablenotes}
                \footnotesize
                \item[+] A: Annular\_0.09231251\_50\_1;   B: Circular\_0.1436665\_0\_1;\\ C: Bull’s Eye\_0.1436665\_0\_1.2;   D: Circular\_0.09231251\_0\_1; \\
                E: Circular\_0.1436665\_0\_1.2.
            \end{tablenotes}
}
        \end{threeparttable}
    }
\label{tab:LRC}
\end{table}

\minisection{Detection Results}
Here we evaluate the performance of the detector on DRC, MRC, and LRC violation detection tasks, with comparison results presented in \Cref{XRC-comp}. Unitho consistently outperforms existing methods across all tasks. Taking the DRC task as an example, Unitho outperforms the SOTA method arXiv'24 by 0.81\% in F$_1$ score, reduces false alarms by 45\% (from 305 to 166), and accelerates inference by over 19× (from 28.55 ms to 1.47 ms). Although arXiv'24 is based on YOLOv8, its inference process is dominated by the time consumed in PCA generation. This demonstrates our detection component achieves superior performance, while maintaining high prediction accuracy, significantly reduces FA, and locates violation regions with exceptional speed.

\begin{table}[tb]
    \centering
    \caption{Performance comparison of different methods on DRC/MRC/LRC detection tasks (\%).}
    \resizebox{\linewidth}{!}
    {
        \begin{threeparttable}
            {
                \begin{tabular}{l l c c c c c c}
                    \toprule
                    \multicolumn{1}{l}{}
                    Task & Method & Recall & Precision & $F_1$ & \#FA  & AP\tnote{+} & RT (ms) \\
                    \midrule
                    
                    \multirow{4}[1]{*}{\textbf{DRC}}
                    & TCAD’22~\cite{chen2022faster} & 76.81 & 81.58 & 79.12 & 1467 & 78.81 & 596.20 \\
                    & TCAD’24~\cite{shao2024lithohod} & 89.74 & 97.00 & 93.23 & 235 & 88.40 & 39.62 \\
                    & arxiv’24~\cite{wu2024enhancing} & \textbf{98.02} & 96.39 & 97.20 & 305 & 97.68 & 28.55 \\
                    \cmidrule{2-8}    &\textbf{Unitho}& 97.98 &  \textbf{98.04} & \textbf{98.01} & \textbf{166} & \textbf{98.72} & \textbf{1.47} \\
                    \midrule

                    \multirow{4}[1]{*}{\textbf{MRC}}
                    & TCAD’22~\cite{chen2022faster} & 9.14 & 2.40 & 3.81 & 58833 & 2.60 & 581.61 \\
                    & TCAD’24~\cite{shao2024lithohod} & 1.77 & 15.77 & 3.18 & 1496 & 26.30 & 53.84 \\
                    & arxiv’24~\cite{wu2024enhancing} & \textbf{96.32} & 69.53 & 80.76 & 4829 & 80.76 & 31.23 \\
                    \cmidrule{2-8}      &\textbf{Unitho}& 94.68 &  \textbf{92.02} & \textbf{93.33} & \textbf{1301} & \textbf{94.84} & \textbf{1.54} \\
                    \midrule

                    \multirow{4}[1]{*}{\textbf{LRC}}
                    & TCAD’22~\cite{chen2022faster} & 25.97 & 36.26 & 30.27 & 19334 & 10.79 & 593.73\\
                    & TCAD’24~\cite{shao2024lithohod} & 79.38 & 46.18 & 58.39 & 39173 & 65.50 & 54.46 \\
                    & arxiv’24~\cite{wu2024enhancing} & 71.67 & 77.42 & 74.43 & 9563 & 77.53 & 28.52\\             \cmidrule{2-8}      &\textbf{Unitho}& \textbf{79.85} &  \textbf{81.44} & \textbf{80.64} & \textbf{7704} & \textbf{82.24} & \textbf{3.45} \\
                    
                    \bottomrule
                \end{tabular}
            \begin{tablenotes}
                \footnotesize
                \item[+] IoU=0.5 
            \end{tablenotes}
            }
        \end{threeparttable}
    }
    \label{XRC-comp}
\end{table}

\begin{figure}[b]
    \centering
    \includegraphics[width=\linewidth]{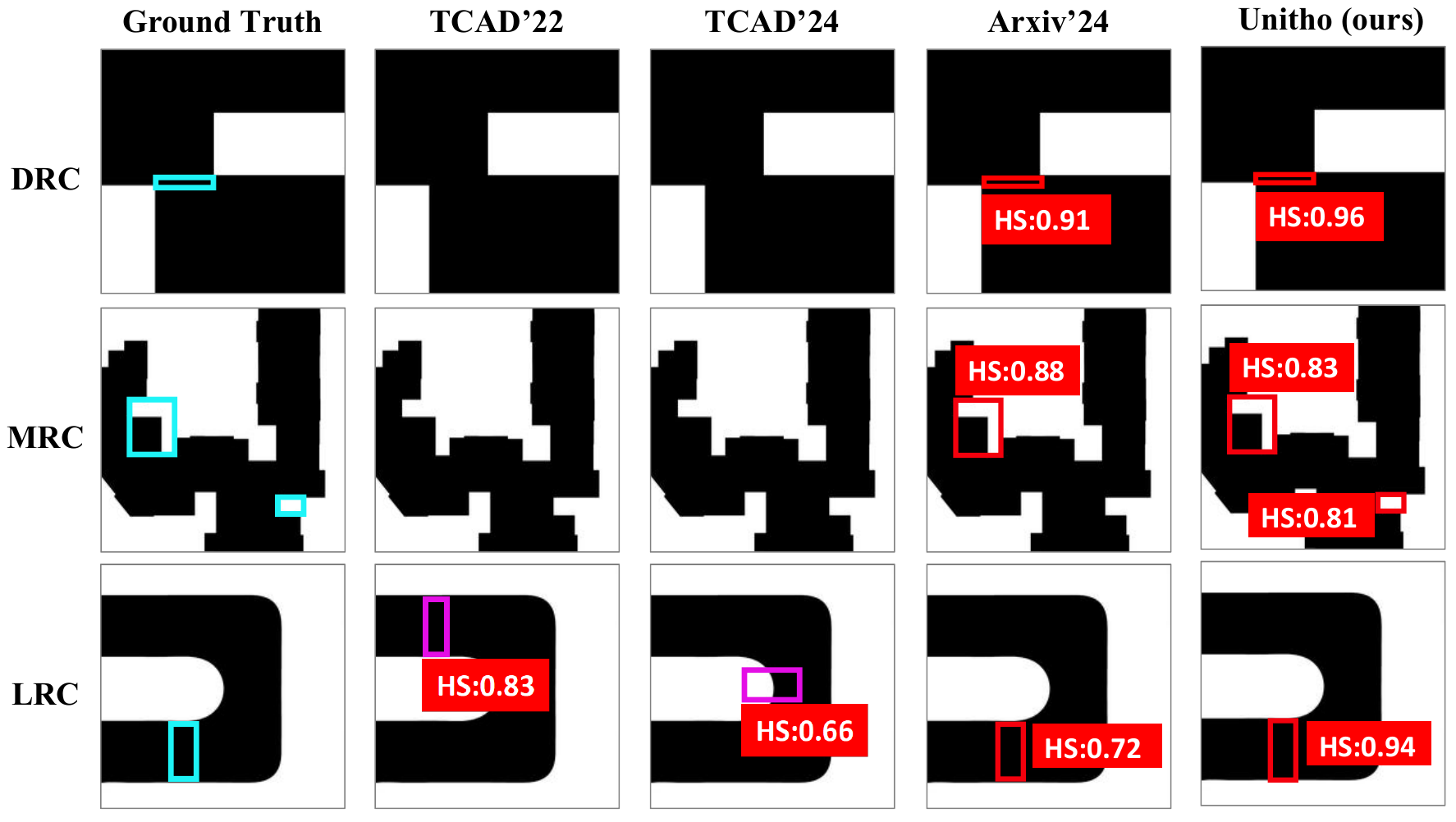}
    \caption{Comparison of detection results. }
    \label{fig:hd}
\end{figure}

\Cref{fig:hd} compares Unitho with baseline methods. Correctly predicted regions are marked with red boxes, missed detection regions with blue boxes, and false positive regions with purple boxes. While others display frequent blue and purple boxes, Unitho predominantly shows red boxes, reflecting higher precision and fewer errors. This confirms Unitho’s effectiveness in localizing hotspots accurately, thereby supporting its practical utility in lithographic hotspot detection.

\subsection{Ablation Study}

We first assess the effect of each sub-component within the reconstruction loss 
As shown in \Cref{tab:gloss}, ``w/o BCE'' denotes the removal of the BCE loss. 
The results show that removing the BCE loss leads to a significant degradation in the reconstruction of local details.
And Dice loss's absence results in a more severe performance drop, highlighting its importance in preserving shape completeness.
The edge-aware loss further enhances edge sensitivity, particularly for mask shapes that exhibit complex, compact structures. After removing this module, mPA and mIoU remain high, but the $F_1$-score drops sharply, indicating poor detail reconstruction.

\begin{table}[tb]
    \centering
    \caption{Ablation studies on reconstruction loss.}
    \resizebox{\linewidth}{!}
    {
        \begin{threeparttable}
            {
                \begin{tabular}{c c c c c c c}
                    \toprule
                    \multirow{2}{*}{Methods} & \multicolumn{3}{c}{Mask Generation (\%)} & \multicolumn{3}{c}{Contour Generation (\%)} \\
                    \cmidrule(r){2-4} \cmidrule(l){5-7}
                    & mPA ↑ & mIoU ↑ & $F_1$ ↑ & mPA ↑ & mIoU ↑ & $F_1$ ↑ \\
                    \midrule
                    w/o BCE  & 91.58 & 88.74 & 90.67 & 94.21 & 90.06 & 91.83     \\
                    w/o Dict  & 90.41 & 86.02 & 87.36 & 93.96 & 89.78 & 91.00     \\
                    w/o Edge & 95.23 & 94.72 &  63.21 & 96.61 & 96.48 & 89.25    \\
                    \midrule
                    \textbf{Unitho (ours)}    & \textbf{98.96}  &  \textbf{96.98} & \textbf{99.08}  & \textbf{99.25}  & \textbf{99.23}   &  \textbf{99.76}         \\ 
                    \bottomrule
                \end{tabular}
            }
        \end{threeparttable}
    }
    \label{tab:gloss}
\end{table}

\begin{table}[tb]
    \centering
    \caption{Ablation studies on contrastive learning objective.}
    \resizebox{\linewidth}{!}
    {
        \begin{threeparttable}
            {
                \begin{tabular}{c c c c c c c}
                    \toprule
                    \multirow{2}{*}{Methods} & \multicolumn{3}{c}{Mask Generation (\%)} & \multicolumn{3}{c}{Contour Generation (\%)} \\
                    \cmidrule(r){2-4} \cmidrule(l){5-7}
                    & mPA ↑ & mIoU ↑ & $F_1$ ↑ & mPA ↑ & mIoU ↑ & $F_1$ ↑ \\
                    \midrule
                    w/o SAC & 94.16 & 87.61 & 89.72 & 95.18 & 89.46 & 88.33    \\
                    w/o PAC & 89.57 & 83.42 & 80.19 & 91.54 & 84.37 & 85.05     \\
                    w/o SAC \& PAC & 80.80 & 72.57 & 75.11 & 85.17 & 80.33 & 81.19     \\
                    \midrule
                    \textbf{Unitho (ours)}    & \textbf{98.96}  &  \textbf{96.98} & \textbf{99.08}  & \textbf{99.25}  & \textbf{99.23}   &  \textbf{99.76}         \\ 
                    \bottomrule
                \end{tabular}
            }
        \end{threeparttable}
    }
    \label{tab:Contrastive}
\end{table}

Next, we evaluated the impact of contrastive learning strategies, as shown in \Cref{tab:Contrastive}. Structure-Aware Contrast (SAC) separates generated images from different layouts, while Process-Aware Contrast (PAC) addresses deviations caused by varying processes within the same layout.
Due to inherent morphological differences across layouts, dropping SAC damages less than dropping PAC, yet either removal noticeably degrades quality. Removing both causes a sharp drop in all metrics, underscoring contrastive learning’s role in process sensitivity, structural discrimination, and generalization.

%% file: doc/conclu.tex
\section{Conclusions}
\label{sec:conclusions}

We propose Unitho, a unified multi-task framework designed to address computational lithography challenges by integrating mask generation, lithography simulation, and violation detection. By fusing layout features with process conditions via a cross-attention fusion module and contrastive learning to sharpen structural and process awareness, Unitho achieves substantial improvements in accuracy and computational efficiency. Evaluations on large-scale industrial datasets reveal that Unitho surpasses existing methods, enabling rapid and precise violation detection, streamlining design workflows, and enabling early co-optimization of layout and process for advanced semiconductor development.